\documentclass[12pt]{article}
\usepackage{judig}
\usepackage[english]{babel}
\usepackage{url}   % simple URL typesetting
\usepackage[hang,flushmargin]{footmisc} % no indentation in footnotes
\usepackage{hyperref}
\usepackage{booktabs}       % professional-quality tables
\usepackage{amsfonts}       % blackboard math symbols
\usepackage{nicefrac}       % compact symbols for 1/2, etc.
\usepackage{xcolor}
\usepackage{graphicx}
\usepackage{tabularx}

\usepackage[]{gb4e}  %for glossed examples
\usepackage{fontspec}
\usepackage{enumitem}
\graphicspath{{img/}}
\title{Wiki Dumps to Training Corpora: \\South Slavic Case}

\paperInfo{Scientific paper}{DOI: 10.xxxx/xxxxx}

\author{
    Mihailo Škorić\textsuperscript{1}\orcidlink{0000-0003-4811-8692}, Cosimo Palma\textsuperscript{2}\orcidlink{0000-0002-8161-9782}
}
\institutions{
 \textsuperscript{1} University of Belgrade, Faculty of Mining and Geology, mihailo.skoric@rgf.bg.ac.rs \par
  \textsuperscript{2} University of Pisa, Department of Computer Sciences, cosimo.palma@phd.unipi.it\par
}

\begin{document}
\maketitle

% In Serbian version, replace "Abstract" with: "Apstrakt" or "Апстракт"
% and "Keywords" with: "Ključne reči" or "Кључне речи"
% If written in another South Slavic language, translate accordingly
\abstractSummaryLang{Abstract}{
This paper presents a pipeline designed to transform raw Wikimedia dumps into quality textual corpora for seven South Slavic languages. The work is divided into two major phases. The first involves extracting and cleaning text from raw dumps of Wikipedia, Wikisource, Wikibooks, Wikinews, and Wikiquote. This step requires careful handling of raw wiki markup to isolate, first of all, textual articles, and then usable natural language text within them. The second phase addresses the challenge of questionable or low-quality articles, which are often generated from databases or structured knowledge bases. These articles are characterised by repetitive patterns, generic phrasing, and minimal to no original content. To mitigate their impact, a n-gram-based filtering strategy was employed to detect high levels of textual redundancy between articles and then remove such articles from the corpora entirely. The resulting datasets aim to provide linguistically rich texts suitable for training language models or conducting comparative research across South Slavic languages. By combining systematic extraction with quality control, this work contributes to the creation of reliable, high-information corpora that reflect the authentic cultural contexts of languages. While focused on the South Slavic case in the paper, the approach is mostly language-agnostic and can be generalised to other languages.}
\keywordslang{Keywords}{Text corpora, Wikimedia projects, Data cleaning}

\section{Introduction}

South Slavic languages (such as Serbian, Croatian, Slovenian and Bulgarian) are notably underrepresented in large-scale natural language processing (NLP) resources compared to other major European languages (such as English, French and German).

Additionally there is a disbalance among the Slavic languages as well, with Russian, Czech and Polish being comparatively well-served. For instance, the Universal Dependencies treebank provides roughly two million training words for Czech and over a million for Russian, but only about 950 words for Ukrainian and none at all for Belarusian, Bosnian, or
Macedonian. 
This issue limits their presence in the multilingual training data used for training of large language models and, more important, limits the development of specialised language-specific models in the region.

To mitigate,
\citep{sharoff-2017-toward} proposes a \emph{Language Adaptation} approach,
analogous to domain adaptation, in which a model for a given NLP task is
built by lexical transfer of cognate words and by learning a new feature
representation for a lesser-resourced (recipient) Slavic language starting
from a better-resourced (donor) one, however, substantial training resources are still needed.

Wikimedia projects offer multiple dumps for South Slavic languages including six national Wikipedia~\citep{wikipedians2004wikipedia}, Wikisource and Wikibooks~\citep{armstrong2010rich} instances, five national Wikiquote instances, three Wikinews~\citep{thorsen2008journalistic} instances and one sole Wikipedia for the Serbo-Croatian macro-language. These projects, with Wikipedia as the richest, offer a valuable source of openly available text, however, transforming raw dumps into usable corpora is not a trivial task~\citep{pasternack2008wikipedia}. Raw wiki markup (wikitext\footnote{\url{https://en.wikipedia.org/wiki/Help:Wikitext}}, templates, and other metadata must be carefully detected via procedural parsing~\citep{dohrn2011design} and/or regular expressions, and stripped (or removed including inner content, depending on the case) to isolate the natural language contents.

In practice, tools such as \textit{WikiExtractor}~\citep{attardi2012wikiextractor} and \textit{mwparserfromhell}\footnote{\url{https://github.com/earwig/mwparserfromhell}} have become standard for handling MediaWiki markup, enabling researchers to systematically strip templates, links, and metadata while preserving the underlying textual content~\citep{song2021research}. 
Another widely adopted practical solution for corpus creation is Wikicorpus module, which is part of a topic modeling and word embeddings library \textit{gensim}~\citep{rehurek-sojka-2010}. The module provides a streamlined interface for downloading and processing Wikipedia XML dumps, automatically stripping markup, normalizing text, and tokenizing sentences. It is designed for scalability, allowing researchers to build large training corpora for distributional semantics and embedding models with minimal preprocessing effort. 

Combining these solutions with language specific processing strategies ensures that corpora derived from Wikimedia projects are reliable and linguistically rich, providing a solid foundation for language model training as well as conducting comparative research across languages.
Such approaches have been applied earlier in other projects for South Slavic languages, for example CLASSLA, which integrates Wikipedia and web crawls into annotated corpora~\citep{markoski-etal-2021-cultural}. 
This is, however, the first effort in the region to include a complete set of Wikimedia projects (Wikipedia, Wikisource, Wikibooks, Wikiquote and Wikinews), while also tackling the detection and removal of a number of potentially templated and/or automatically generated articles. Such can lead to text redundancy, reduced linguistic diversity and word frequency distributions disproportionate to the actual language, all of which can distort statistical distributions during potential language model training, so its removal is highly desirable.

The paper is organised into four main sections including this one. The \textit{Data} section describes the sources used, describes the fetching of dump files from the web and their transformation into readable format and detailing the availability of Wikimedia projects across languages (e.g., Macedonian Wikinews, Slovenian Wikiquote, Serbo-Croatian Wikipedia) The following section, \textit{Methodology}, is describing the process of corpora compilation and is divided into two subsections: \textit{Text Extraction}, which explains the technical pipeline for parsing the dumps, cleaning markup, and isolating natural language text using \textit{mwparserfromhell} in combination with various regular expressions and text-processing functions; and \textit{Filtering}, which focuses on detecting and removing templated articles through deduplication heuristics, ensuring a proper word-frequency distribution. Finally, the \textit{Discussion} evaluates the resulting corpora, comparing statistics across languages, reflecting on the impact of filtering, and situating the contribution within broader multilingual corpus creation efforts.

\section{Data}

All corpora are derived from Wikimedia project dumps dated April 1\textsuperscript{st} 2026. (version code \texttt{20260401}).

The process begins with retrieving the raw Wikimedia dump files in compressed form and preparing them for analysis. Each dump is downloaded directly from the official Wikimedia servers\footnote{\url{https://dumps.wikimedia.org}}, in its original \textit{.xml.bz2} format, while warranting version consistency across languages and projects. Once obtained, the compressed archives are decompressed and parsed into a structured format suitable for further processing.

During parsing, we iterate through the XML structure of the dump, identifying individual page entries. Only articles from the main namespace (articles) are retained, while redirects, administrative pages, and other non-content entries are excluded. For each valid article, the title, identifier, and textual content are extracted. To avoid noise from near-empty pages, a minimum text length threshold (80 characters) is applied, ensuring that only substantive articles are preserved.

The extracted articles are then serialised into a line-oriented JSON format (JSONL), where each line corresponds to a single page with its metadata and raw text. This representation provides a lightweight and easily processable structure for subsequent cleaning and filtering steps. By standardizing the output across all languages and projects, the procedure guarantees comparability and facilitates downstream corpus construction.

Table~\ref{tab:datasets} lists the available sets by language with a total number of articles per existing set.

\begin{table}[h]
\centering
\caption{Wikimedia datasets included in the experiment (dump date: April 1\textsuperscript{st} 2026). The rows indicate different languages (represented by respective ISO 639-1 language codes), columns different Wikimedia projects, and the numbers a total of pages per set. } 
\begin{tabularx}{\textwidth}{X|c|c|c|c|c}
\toprule
\rowcolor{headergray}
 & Wikipedia & Wikisource & Wikiquote & Wikibooks  & Wikinews \\
\midrule
sr      & 712,843 & 43,115 & 6,049 & 1,680 & 53,141 \\ \midrule
hr    & 230,550 & 12,119 & 2,328 & 1,216 & -- \\ \midrule
bs       & 97,437 & 1,889 & 4,446 & 53 & 367 \\ \midrule
sl     & 197,089 & 24,991 & 3,142 & 348 & -- \\ \midrule
mk    & 160,051 & 3,740     & -- & 122 & -- \\ \midrule
bg     & 308,972 & 2,517 & 4,756 & 555 & 430 \\ \midrule
sh  & 461,454 & --      & --      & --      & -- \\ 
\bottomrule

\end{tabularx}

\label{tab:datasets}
\end{table}

The data summarised in Table~\ref{tab:datasets} shows an unevenness of Wikimedia project activity across the South Slavic languages. Serbian stands out with the most numerous Wikipedia corpus, exceeding 700,000 articles, and also maintains substantial Wikisource and Wikinews collections. Croatian and Slovenian have mid-sized Wikipedias, accompanied by moderate contributions in Wikisource and Wikiquote, while Bosnian and Macedonian projects are smaller in scale, with only a few thousand articles in their secondary projects. Finally, Bulgarian shows a relatively large Wikipedia and balanced representation across other projects. The Serbo-Croatian has only Wikipedia data available, with none of the other sister projects active.

\section{Methodology}

\subsection{Text extraction}
\begin{figure}[htbp]
    \centering
    \includegraphics[width=\linewidth]{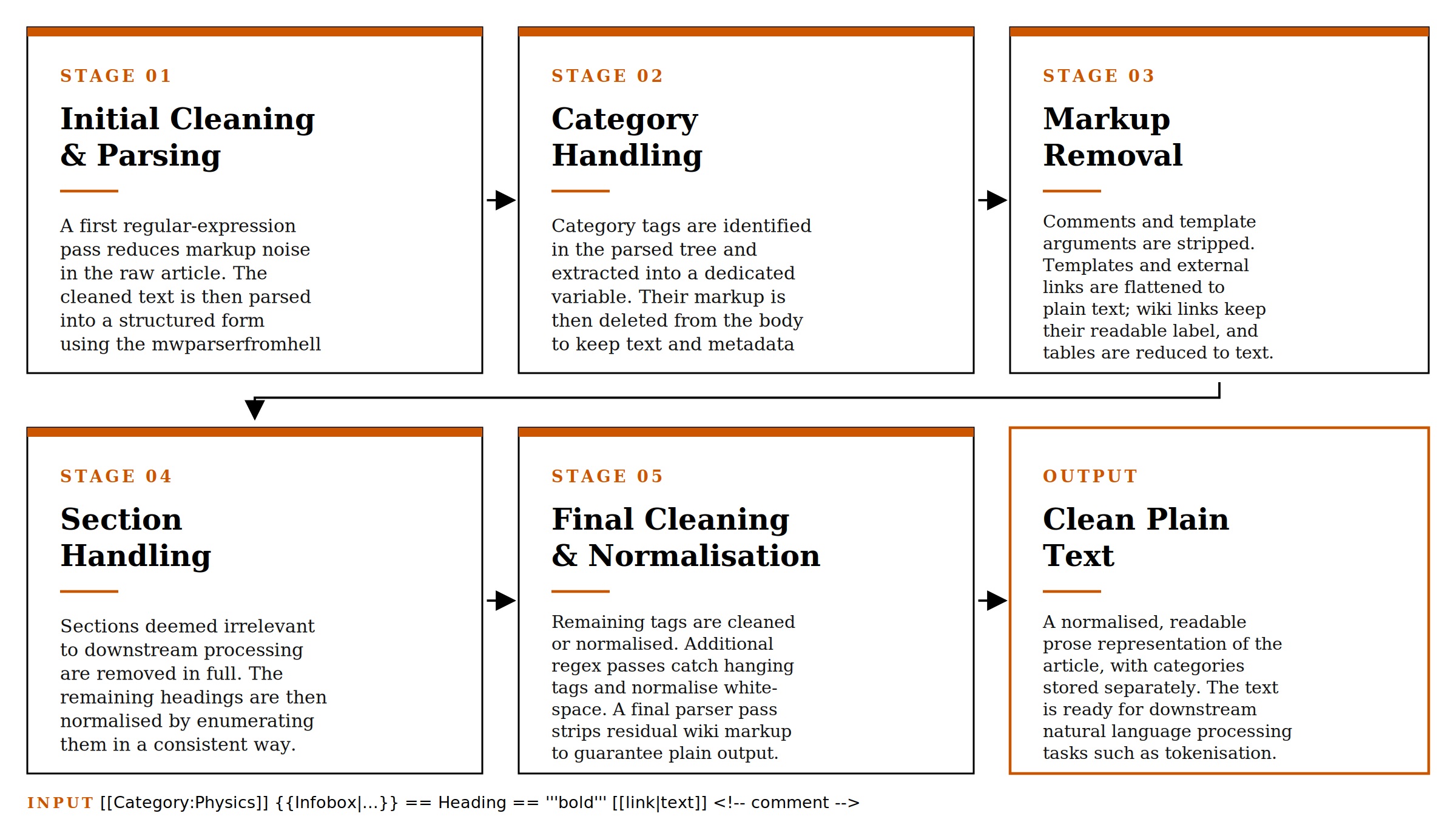}
    \caption{Wikipedia text preprocessing pipeline. A five-stage transformation
    from raw MediaWiki markup to clean plain text.}
    \label{fig:textpreprocessing}
\end{figure}

Once the raw dumps are converted into JSONL files, each page is processed in batches to extract usable text and metadata. The procedure distributes work across multiple processes to handle large volumes efficiently, while monitoring for timeouts or errors to ensure robustness. For every article, the text is cleaned and enriched with additional information such as a canonical URL, word counts and percentage of Cyrillic words. Valid results are written back into a structured JSONL file, and summary statistics on the number of articles and total word count are recorded. This step transforms the initial parsed dumps into a standardised, quality-controlled dataset that is ready for deeper linguistic filtering. The process is performed over five general steps (Figure~\ref{fig:textpreprocessing}).

The first cleaning stage applies a sequence of regular expression--based transformations designed to strip away the most common forms of wiki markup and structural noise. 
The preprocessing pipeline transforms raw MediaWiki source into clean plain text through five sequential stages. In the first stage, an initial regular-expression pass reduces markup noise, after which the text is parsed into a structured representation using the \texttt{mwparserfromhell} library. The second stage handles categories: category tags are identified and extracted into a separate variable, and the corresponding markup is removed from the body. The third stage performs the bulk of the markup removal. Comments and template arguments are stripped; templates and other constructs are flattened into plain text; residual wiki templates and external link markup are deleted; wiki links are processed to retain only their readable text; and tables are reduced to their textual content. The fourth stage operates at the section level, removing unwanted sections from the article and normalising the remaining headings by enumerating them consistently. In the final stage, tag cleaning removes or normalises any remaining markup, additional regex passes catch hanging tags and normalise whitespace, and a last parser pass strips residual wiki constructs to guarantee a plain-text output.

Each step will be explained in more detail in the following sections.

\subsubsection{Initial cleaning and parsing}

This initial regex pass reduces markup complexity, leaving behind text that is closer to natural prose. By removing templates, media references, and structural tags early, subsequent parsing stages can operate on cleaner input without being overwhelmed by nested or malformed constructs resulting in much faster parsing using the \texttt{mwparserfromhell} library, which transforms wikitext into a node-like structure for the following processing steps.

\subsubsection{Category Handling}

The next step focuses on identifying and isolating category information embedded in the wiki markup. Each page is scanned for wikilinks nodes, and those that begin with a category prefix are recognised as category tags. The category name is extracted by removing the prefix (e.g.\ \texttt{Category:History} becomes simply \texttt{History}), and stored separately into a list. At the same time, the original category markup is removed from the text. 

This procedure ensures that the cleaned text remains free of non-prose elements, while the category information is preserved in a structured form. The resulting output therefore consists of plain text of the article and a list of associated categories for each article. This separation allows the corpora to retain valuable metadata for downstream tasks such as genre classification or thematic filtering, without compromising the integrity of the textual content itself.

\subsubsection{Markup Removal}
%\subsubsection{Comments, Arguments, and Template Processing}

Following category extraction, the text is further refined by removing or simplifying nodes that do not contribute to the linguistic content of the corpora. Embedded comment nodes enclosed in \texttt{<!--~...~-->}, which typically hold editorial notes or hidden instructions rather than usable text, are removed entirely, as are the argument markers and placeholders carried by certain templates, which have no meaning outside the wiki environment. Templates themselves are the central feature of Wikimedia markup and require more careful treatment: the pipeline identifies all template nodes and sorts them by length so that larger, more complex templates are handled first. Templates belonging to a predefined \textit{keep list} (e.g.\ \textit{ppoem} and \textit{cquote}) are preserved in simplified form, with their parameters extracted and concatenated into plain text; all others are discarded so that only linguistically relevant material remains. Finally, a broader sweep removes any template structures enclosed in double braces (\texttt{\{\{~...~\}\}}) that have survived the previous pass, by scanning for opening braces and matching them with their corresponding closing braces even in cases of nesting, yielding a clean removal of markup ranges while leaving the surrounding text intact.

Together, these steps eliminate hidden comments, argument placeholders, and non-essential templates while retaining the textual content of those templates deemed linguistically valuable, so that the corpus reflects the text rather than structural or formatting artefacts.

%\subsubsection{Wikilinks and Tables}

The next stage of cleaning addresses two particularly complex sources of markup: wikilinks not captured by the previous regular-expression pass, and tables. All remaining link nodes are inspected and processed: links pointing to images or files are removed entirely, as they do not contribute textual content, while for the rest the visible text is preserved, mirroring the earlier pass. If a link takes the form \texttt{[[Target|Visible]]}, only the \textit{Visible} part is retained; if no alternate text is provided, the link title itself is kept. In this way, the corpus reflects the human-readable surface rather than the underlying link syntax.

Tables are handled separately, since they are a frequent source of markup complexity, often containing nested structures and irregular formatting. The pipeline first balances unclosed table tags by adding missing delimiters where necessary, which is rare but does happen, especially when an article or a section ends with a table. Each table is then parsed row by row, with row markers and delimiters discarded; cell contents are cleaned using regular expressions, and the remaining text is concatenated into a plain representation. Nested or innermost tables are processed first to avoid structural errors. When the parser fails to detect an expected table, greedy regex matching against \texttt{\{|~...~|\}} is applied as a fallback, followed by a secondary match using \texttt{|-~...~==} to capture remaining row elements up to the next section mark. Any stray closing tags are removed to prevent malformed output. The final result is a simplified textual rendering of the table contents, stripped of markup but retaining the information that was encoded in its cells.

By simplifying the remaining links and flattening tables into plain text, this stage removes two of the most markup-heavy structures in Wikimedia dumps. The resulting text is more coherent and readable, while still preserving most of the substantive information originally encoded in links and tabular data.

\subsubsection{Section Handling and Headings}

After links and tables are simplified, the text is further refined by selectively retaining or discarding sections, and by normalising all remaining headings. Because the three operations are logically heterogeneous, with one project (Wikiquote) following inverted logic compared to the rest, they are presented here as a list:

\begin{enumerate}
  \item \textbf{Section removal}:
  For most Wikimedia projects (all but Wikiquote), only the main textual sections are retained. Sections whose headings match a predefined list of unwanted titles (such as \textit{References}, \textit{Gallery} and \textit{External links}, including appropriate localisation variants) are discarded. The procedure also discards empty sections and those consisting solely of a heading, while substantive sections are preserved. This step prevents non-prose material and empty sections from entering the corpus.

  \item \textbf{Section filtering (Wikiquote)}:  
  In the case of Wikiquote, the logic is reversed: only sections explicitly marked as containing quotations (such as \textit{quotes}, \textit{sourced} and \textit{attributed}, including appropriate localisation variants) are retained, while all others are removed. This guarantees that the resulting corpus consists exclusively of the intended content type, that is, quotations themselves, without any descriptions or metadata.

  \item \textbf{Heading processing}:  
  Headings are normalised and enumerated to provide a consistent hierarchical structure. Each heading level is tracked with counters, producing a numbering scheme (e.g.\ \textit{1}, \textit{1.1}, \textit{1.2}) that reflects the document outline. The heading text itself is stripped of markup and reinserted into the text with the corresponding enumeration. This creates a clear, standardised representation of the article's structure, increasing the readability of the text.
\end{enumerate}

By removing unwanted sections or filtering project-specific content, the corpora retain only relevant textual material while preserving a coherent structural outline of each article through the normalisation of the remaining headings.

\subsubsection{Final Cleaning and Normalisation}

The last stage of text extraction applies a series of targeted clean-up operations to remove residual markup and ensure that the output is plain, coherent text. All remaining HTML-like tags are detected via regular expression and inspected. Their treatment depends on which of three categories they fall into, as summarised in Table~\ref{tab:tagpolicy}: tags in the \textit{destroy list} are removed together with their inner content, tags in the \textit{preserve list} are kept intact because they carry meaningful inline information, and any other tag has its markup stripped while its inner text is retained. %This ensures that only linguistically relevant text remains, while structural or decorative tags are discarded.

\begin{table}[h]
\centering
\caption{Treatment of HTML-like tags during final cleaning.}
\label{tab:tagpolicy}
\begin{tabularx}{\textwidth}{l|X|l}
\toprule
\rowcolor{headergray}
Category & Tags & Action \\
\midrule
Destroy list & \textit{noinclude}, \textit{ref}, \textit{gallery}, \textit{timeline} & remove tag and content \\ \midrule
Preserve list & \textit{math}, \textit{code}, \textit{syntaxhighlight}, \textit{b}, \textit{sup}, \textit{sub} & keep tag and content \\ \midrule
Other & any remaining tag & strip tag, keep content \\
\bottomrule
\end{tabularx}
\end{table}

Cross-language interwiki links such as \texttt{[[fr:Page]]} are removed, since they point to external projects rather than contributing text, and incomplete or hanging template fragments such as \texttt{\{\{something|} are collapsed to prevent malformed markup from appearing in the corpus. A secondary parsing pass is then applied to strip any remaining wiki-markup nodes, reducing headings, links, and other constructs to plain text and so guaranteeing that the output is free of syntactic artefacts. Finally, a closing regular-expression sweep removes special \textit{magic words} such as \texttt{\_\_TOC\_\_} (which marks a table of contents), stray closing tags not on the preserve list, and leftover template attributes such as key--value pairs. Additional replacements handle dangling link markers and language-specific constructs, and whitespace is normalised so that multiple spaces, tabs, or newlines are collapsed into a consistent format.

The final cleaning stage removes that all residual wiki markup, structural tags, and formatting artefacts. The article text is left in a standardised form, with normalised spacing and consistent structure, ready for inclusion in the compiled corpora.

% TODO
\subsection{Filtering}

This stage applies the combination of deduplication heuristics and similarity analysis filtering procedures designed to detect and remove mechanically generated texts, increasing the likelihood of authentic, human written ones. The filtering reduces the presence of templated articles, improving the overall corpus authenticity.

Beyond the initial safeguard of discarding texts shorter than eighty characters, filtering proceeds in three main steps. First, each text is encoded into a vector representation, more suitable for further processing. Second, articles are grouped into clusters according to their extracted categories, allowing comparisons within smaller sets, under the assumption that highly similar articles also share categories. Finally, similarities are calculated inside each cluster to detect templated or generated articles, which are then removed.

\subsubsection{Encoding Text into Vectors}

The first step of filtering transforms each article into a numerical representation suitable for processing, through a combined operation of tokenisation, vocabulary construction, and encoding. Each article is first normalised and split into a sequence of tokens: text is lowercased, digits are replaced with placeholders, and the resulting string is segmented into words and symbols. Token counts are then accumulated across the  dataset to build a single vocabulary, with tokens occurring fewer than three times discarded as noise; the remaining tokens are sorted by frequency and assigned unique indices. Beyond serving as the basis for encoding, this vocabulary also provides dataset-level token-frequency statistics that are used later in the pipeline.

Articles exceeding 2{,}000 words are excluded from the similarity check at this point, both to avoid skew from excessively long or anomalous texts and to speed up subsequent pairwise comparisons, under the assumption that longer texts are less likely to be template-generated. For each remaining article, the first 500 tokens are replaced with their respective vocabulary indices to produce a fixed-length vector; alongside the vector, metadata such as the article identifier and its subject categories are preserved. The choice to truncate to the opening of each article rests on the assumption that templated content is most readily identifiable from its beginning. The numerical parameters governing this step are summarised in Table~\ref{tab:encoding}. The encoded vectors are finally written to a JSONL file, forming a structured dataset that can be reloaded efficiently without repeating the encoding process.

\begin{table}[h]
\centering
\caption{Parameters used in the vector-encoding stage.}
\label{tab:encoding}
\begin{tabularx}{\textwidth}{X|r}
\toprule
\rowcolor{headergray}
Parameter & Value \\
\midrule
Minimum token frequency for inclusion in the vocabulary & 3 \\ \midrule
Maximum article length considered for encoding (words) & 2{,}000 \\ \midrule
Number of leading tokens encoded per article (vector length) & 500 \\
\bottomrule
\end{tabularx}
\end{table}

By encoding text into vectors, the corpus is transformed into a format that enables quantitative comparison. Vectors can now be analysed systematically, providing the foundation for similarity detection.

\subsubsection{Clustering by Categories}

Once articles are encoded into vectors, the next step is to organise them into clusters according to their subject categories, so that computationally expensive similarity analysis is performed within thematically coherent groups rather than across unrelated material. Each record is examined for its associated category field, and the corresponding labels are extracted. Since an article may belong to a single or multiple categories, it is indexed under each of them. Articles sharing the same category are then grouped together into buckets, forming initial clusters of topically related texts. To prevent distortion, and to avoid the quadratic blow-up in pairwise comparisons that comes with overly large groups, buckets exceeding a maximum size of 3{,}000 articles are split into smaller chunks of at most the same size.

By clustering articles according to their categories and chunking oversized groups, this step establishes a structured environment for later filtering. Comparisons are restricted to thematically consistent sets, which increases the reliability of similarity detection in the final stage.

\subsubsection{Similarity Analysis within Clusters}

The final filtering step applies deduplication heuristics to detect and remove templated or low-effort articles by measuring their mutual similarity within each cluster.

\begin{enumerate}

  \item\textbf{MinHashing}:

  Traditional Jaccard similarity~\citep{jaccard1901etude} measures the overlap between two sets of $n$-grams, defined as:
  \[J(A,B) = \frac{|A \cap B|}{|A \cup B|}\]
  where $A$ and $B$ are the sets of $n$-grams extracted from two sequences. While exact Jaccard computations are accurate, they are computationally expensive when applied to large clusters of documents such as this one, especially if there are many pairwise comparisons.

  In order to efficiently detect near-duplicate sequences and recurring templates, \textit{MinHashing}~\citep{broder1997resemblance} is adopted as a similarity scoring alternative to approximate the Jaccard similarity. Each set of $n$-grams is hashed multiple times under independent permutations, and the minimum hash value for each permutation is recorded. The resulting signature is a compact representation of the set. The fraction of matching positions between two signatures approximates their Jaccard similarity while drastically reducing computation costs. Instead of comparing hundreds of $n$-grams directly, we compare fixed-length signatures (e.g., 128 integers). This makes MinHash particularly well-suited for large-scale deduplication tasks.

  This method had already been applied in other deduplication systems such as \textit{One instance only} (ONION)~\citep{pomikalek2011removing}, where it demonstrated the effectiveness of $n$-gram based similarity for detecting boilerplate and template reuse in Wikipedia articles. At the moment, this method is widely used in document deduplication, plagiarism detection, and large-scale clustering applications.

  \item\textbf{Similarity Scoring}

  To measure similarity between articles in this case, MinHash signatures built on trigram representations are deployed. Each sequence is first decomposed into contiguous trigrams, which are then hashed under multiple permutations, and the minimum hash values are recorded to form a compact, fixed-length signature. 

  Within a cluster of records, pairwise similarities are computed by comparing these hash signatures and the similarity score is defined as the fraction of matching positions between two signatures. Pairs exceeding similarity threshold of $0.5$ are noted, and for each record the top three highest similarity scores across all clusters are saved.

  \item\textbf{Cutoff}
  For each article we calculate a single score as the average of previously saved top three scores. If there are fewer than three scores for an article, zero-padding is performed before calculating the average. 
  
  Once there is a score for each article, the scores are compiled into a single sorted list and evaluated using the \texttt{KneeLocator} algorithm~\citep{satopa2011kneedle} in order to determine the cutoff point. Namely, the point where the similarity distribution changes most sharply (knee of the curve), is presumed to be separator between template-generated and human-written content. Thus, all articles with scores higher than the cutoff point are eliminated from the corpus.

\end{enumerate}

Table~\ref{tab:wikicleaning} compares the number of articles and words across selected South Slavic and Balkan Wikipedias before and after the filtering stage, which results in substantial reductions in article and word counts  for several languages.

\begin{table}[h]
\centering
\caption{Number of Wikipedia articles and words before and after the filtering stage for each language.}
\label{tab:wikicleaning}
\begin{tabularx}{\textwidth}{X|r|r|r|r}
\toprule
\rowcolor{headergray}
& Articles Before & Articles After & Words Before &  Words After \\
\midrule
sr & 528,932 & 224,439 & 354,948,022 & 134,239,786 \\ \midrule
bg & 243,675 & 230,433 & 101,125,608 & 99,266,154 \\ \midrule
sl & 137,994 & 119,566 & 70,351,709  & 68,069,867  \\ \midrule
hr & 151,761 & 140,346 & 59,807,937  & 57,903,367  \\ \midrule
mk & 112,888 & 93,274 & 57,632,973  & 52,382,161  \\ \midrule
bs & 76,426  & 46,472  & 29,904,610  & 24,372,685  \\ \midrule
sh & 221,523 & 121,985 & 58,458,749  & 49,873,695  \\ 
\bottomrule
\end{tabularx}
\end{table}

\section{Discussion}

This paper had been focused on the extraction, cleaning, and filtering of textual data from Wikimedia projects in seven South Slavic languages. The methodology combined markup stripping and similarity analysis to improve the probability that the resulting corpora consist of authentic, naturally written texts.

\subsection{Extraction results}

The results of this experiment are summarised in Table~\ref{tab:datasets2}, which presents the count of extracted words for different Wikimedia projects and for each language. It shows the contributions of Wikipedia, Wikisource, Wikiquote, Wikibooks, and Wikinews, and provides insight into the diversity and balance of available resources for each of the selected languages.

\begin{table}[h]
\centering
\caption{Wikimedia datasets produced by the experiment (dump date: April 1\textsuperscript{st} 2026). Rows indicate languages, columns indicate different Wikimedia projects, and numbers represent total words per set.}
\label{tab:datasets2}
\begin{tabularx}{\textwidth}{X|r|r|r|r|r}
\toprule
\rowcolor{headergray}
 & Wikipedia & Wikisource & Wikiquote & Wikibooks & Wikinews \\
\midrule
sr & 134,239,786 & 21,829,353 & 1,007,368 & 695,980 & 7,544,986 \\ \midrule
bg & 99,266,154 & 3,185,352  & 7,360     & 381,109 & 120,270 \\ \midrule
sl & 68,069,867  & 61,444,181 & 233,132   & 627,769 & -- \\ \midrule
hr & 57,903,367  & 9,680,310  & 5,001     & 299,653 & -- \\ \midrule
mk & 52,382,161  & 1,149,472  & --        & 32,339  & -- \\ \midrule
bs & 24,372,685  & 2,271,470  & 143,188   & 29,618  & 45,485 \\ \midrule

sh & 49,873,695  & --         & --        & --      & -- \\ 
\bottomrule
\end{tabularx}
\end{table}

The distribution of extracted words highlights important differences in how each language community contributes to the ecosystem:

\begin{itemize}
  \item \textbf{Serbian (sr)} dominates overall, with over 134 million words from Wikipedia alone. It also has substantial contribution from Wikisource, which is second largest (21.8 million words). Its Wikinews (7.5 million words) and Wikiquote (1 million words) projects are the only substantial projects of those categories in the region. All of this makes it the most rich and diverse dataset in the region in terms of project coverage, with the highest word count in four out of five projects.

  \item \textbf{Bulgarian (bg)} stands out with the second largest Wikipedia dataset, nearing 100 million words. In comparison, its Wikisource seems small with 3.2 million words. It is also noted that it has all five projects active.

  \item \textbf{Slovenian (sl)} stands out for its unusually large (largest) Wikisource dataset (61.4 million words), nearly equal to its Wikipedia size (68 million). It also boasts the second largest Wikibooks. This reflects a strong emphasis on digitized literary and historical texts. It does not have a Wikinews project, but is still the second largest set overall.
  
  \item \textbf{Croatian (hr)} is moslty concentrated in Wikipedia (57.9 million words), with a notable secondary contribution from Wikisource (9.7 million). Same as Slovenian, there is no active Wikinews project. Other projects remain marginal.

  \item \textbf{Macedonian (mk)} is primarily Wikipedia-driven (52.4 million words), with a modest contribution to Wikisource and Wikibooks projects. There are no active Wikiquote and Wikinews projects for Macedonian.
  
  \item \textbf{Bosnian (bs)} has the smallest total dataset among independent language projects, but all five projects active. The largest portions of text are coming from Wikipedia (24.4) and Wikisource (2.3 million). The contributions to Wikiquote, Wikibooks and Wikinews is marginal.
  
  \item \textbf{Serbo-Croatian (sh)} is represented only by Wikipedia with nearly 50 million words.
\end{itemize}

\subsection{Filtering results}

\begin{figure}[ht]
\centering
    \includegraphics[width=\linewidth]{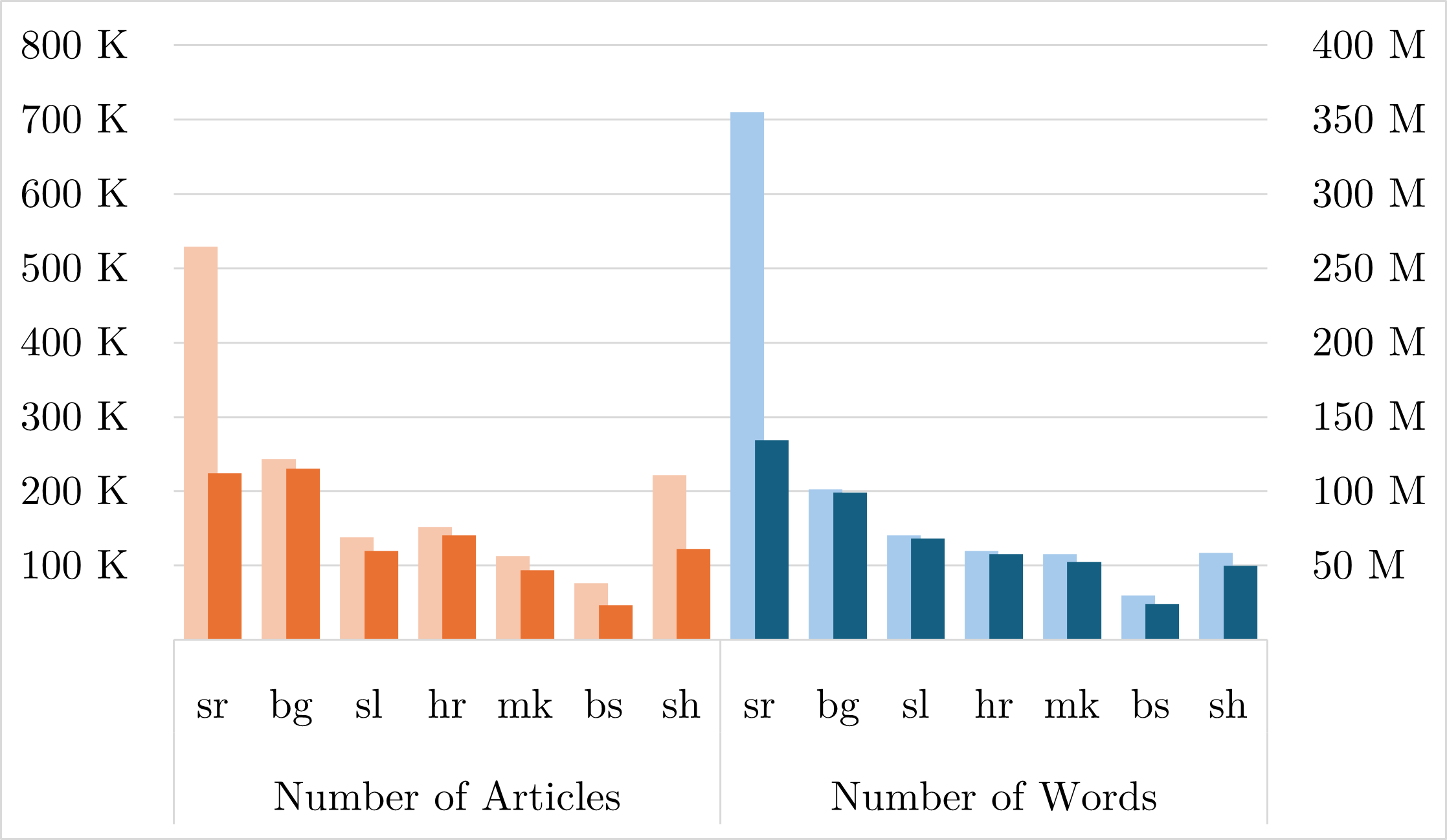}
\caption{Comparison of article and word counts before and after the filtering stage across all languages.}
\label{fig:img1}
\end{figure}

The second subject that should be discussed are the results of the Wikipedia article filtering presented in Table~\ref{tab:wikicleaning} and visualised here in Figure~\ref{fig:img1}.

After applying the similarity scoring and cutoff procedure, the size of the corpus was substantially reduced while its diversity was retained. The Serbian corpus (\texttt{sr}) shows the largest reduction, losing nearly half of its articles and reflecting the extensive presence of duplicated or templated material in this set. This also greatly affected the word count, which dropped by roughly 200 million words (around 58\%). Other languages such as Bosnian (\texttt{bs}) and Serbo-Croatian (\texttt{sh}) experienced significant decreases as well, indicating high levels of redundancy. In contrast, Bulgarian (\texttt{bg}) and Slovenian (\texttt{sl}) retained most of their articles, suggesting comparatively lower duplication.

In order to further assess the impact of filtering, we compare the word frequencies of a general corpus (1) against the word frequencies in the Wikipedia corpus before (2) and after filtering (3). More precisely, we calculate the cosine delta distance between normalised token-frequency vectors~\citep{cinkova2020stylometry} for the top 100 most frequent tokens of SrpKor2013~\citep{vitas2024many} (1) and of the Serbian Wikipedia corpus before (2) and after filtering (3). The results, presented in Table~\ref{tab:similarity}, show that the filtering pushed the token distribution of the Wikipedia set toward that of a more general corpus. While the distance is still smallest between the two Wikipedia corpora, the distances between each of them and the general corpus paint a clear picture.

\begin{table}[h]
\centering
\begin{tabularx}{\textwidth}{X|r|r|r}
\toprule
\rowcolor{headergray}
 & Wikipedia before & Wikipedia after & SrpKor2013 \\
\midrule
Wikipedia before & 0 & 0.2158 & 0.7181 \\ \midrule
Wikipedia after & 0.2158 & 0 & 0.2716 \\ \midrule
SrpKor2013    & 0.7181 & 0.2716 & 0 \\
\bottomrule
\end{tabularx}
\caption{Cosine delta distance scores between token-frequency vectors of three different corpora.}
\label{tab:similarity}
\end{table}

Overall, the filtering step eliminated millions of redundant words across the corpora, demonstrating the effectiveness of the filtering pipeline. Despite the radical filtering of Wikipedia articles, the Serbian set remained the largest by a clear margin, highlighting the scale of that edition and the persistence of unique material even after aggressive pruning.

The comparative results also highlight important differences across languages: while Bulgarian and Slovenian corpora retained most of their articles, Bosnian and Serbo-Croatian experienced sharper reductions, reflecting higher levels of redundancy. This variation suggests that duplication patterns are not uniform across editions, but instead shaped by editorial practices and community size.

%By removing repetitive structures while preserving distinctive texts, the filtering process improves corpora quality and ensures that subsequent use is based on cleaner, more representative data. In this way, resulting corpora are both leaner and more reliable for downstream research.

\acknowledgmentLang{Acknowledgements}{
This research was supported by the Science Fund of the Republic of Serbia, \#7276, Text Embeddings - Serbian Language Applications - TESLA.
}

\urlstyle{same} 

%\vspace{10pt}
% In Serbian version, replace section title with: Ресурси or Resursi
%\section*{Language Resources}
%\bibliographystylelanguageresource{judig-natbib}
%\bibliographylanguageresource{languageresource}

% In Serbian version, replace section title with: Literatura or Литература
\section*{References} 
\bibliographystyle{judig-natbib}
\bibliography{references}

\newpage
\titlelang{Od Viki izvoza \\do korpusa za obučavanje:\\ Slučaj južnoslovenskih jezika}
%\savedAuthors
{\raggedright Mihailo Škorić, Cosimo Palma\par}

% If the summary is supposed to be in English, use "Summary" instead of "Sažetak"
% and "Keywords" instead of "Ključne reči".
\abstractSummaryLang{Sažetak}{
Ovaj rad predstavlja proces osmišljen da transformiše sirove Vikimedijine dampove u kvalitetne tekstualne korpuse za sedam južnoslovenskih jezika. Rad je podeljen u dve glavne faze. Prva uključuje ekstrakciju i čišćenje teksta iz sirovih izvoza Vikipedije, Vikizvornika, Vikiknjiga, Vikivesti i Vikicitata. Ovaj korak zahteva pažljivo rukovanje sirovim viki oznakama kako bi se izolovali, pre svega, tekstualni članci, a zatim i upotrebljiv tekst na prirodnom jeziku unutar njih. Druga faza se bavi izazovom sumnjivih ili nekvalitetnih članaka, koji se često generišu iz baza podataka ili strukturiranih baza znanja. Ove članke karakterišu ponavljajući obrasci, generičko fraziranje i minimalan ili nikakav originalni sadržaj. Da bi se ublažio njihov uticaj, korišćena je strategija filtriranja zasnovana na n-gramima kako bi se otkrili visoki nivoi tekstualne redundantnosti među člancima, pri čemu se detektovani članci u potpunosti uklonjaju iz korpusa. Dobijeni skupovi podataka imaju za cilj da obezbede lingvistički bogate tekstove pogodne za obuku jezičkih modela ili sprovođenje komparativnih istraživanja na južnoslovenskim jezicima. Kombinovanjem sistematske ekstrakcije sa kontrolom kvaliteta, ovaj rad doprinosi stvaranju pouzdanih, visokoinformativnih korpusa koji odražavaju autentične kulturne kontekste jezika. Iako se rad fokusira na slučaj južnoslovenskih jezika, pristup je uglavnom jezički nezavistan i može se primeniti na druge jezike i porodice jezika.}

\keywordslang{Ključne reči}{Tekstualni korpusi, projekti Vikimedije, Čišćenje podataka}

\end{document}